\documentclass[]{article}
\usepackage{graphicx}
\usepackage{subfigure}

\long\def\comment#1{}
\newcommand{\commentout}[1]{}

\newcommand{\eg}{{\em e.g.}}
\newcommand{\ie}{{\em i.e.}}

\newcommand{\secref}[1]{Section~\ref{#1}}
\newcommand{\eqref}[1]{Equation~\ref{#1}}
\newcommand{\figref}[1]{Figure~\ref{#1}}

\begin{document}

%\title{Analysis of Dynamic Task Allocation Experiments in Robots}

%\author{Kristina Lerman, Chris V. Jones, Aram Galstyan, Maja J. Matari{\'c}}

%\maketitle

\begin{center}
{\LARGE\bf Analysis of Dynamic Task Allocation in Multi-Robot Systems}\\[0.25cm]
{\bf Kristina Lerman$^1$, Chris Jones$^2$, Aram Galstyan$^1$ and Maja J Matari\'{c}$^2$}\\
\end{center}

\begin{center}
1. Information Sciences Institute \\
2. Computer Science Department \\
University of Southern California, Los Angeles, CA
90089-0781, USA \\
\{lerman\(|\)galstyan\}@isi.edu, \{cvjones\(|\)maja\}@robotics.usc.edu
\end{center}

\begin{abstract}

Dynamic task allocation is an essential requirement for
multi-robot systems operating in unknown dynamic environments. It
allows robots to change their behavior in response to
environmental changes or actions of other robots in order to
improve overall system performance. Emergent coordination
algorithms for task allocation that use only local sensing and no
direct communication between robots are attractive because they
are robust and scalable. However, a lack of formal analysis tools
makes emergent coordination algorithms difficult to design. In
this paper we present a mathematical model of a general dynamic
task allocation mechanism. Robots using this mechanism have to
choose between two types of task, and the goal is to achieve a
desired task division in the absence of explicit communication and
global knowledge. Robots estimate the state of the environment
from repeated local observations and decide which task to choose
based on these observations. We model the robots and observations
as stochastic processes and study the dynamics of the collective
behavior. Specifically, we analyze the effect that the number of
observations and the choice of the decision function have on the
performance of the system. The mathematical models are validated
in a multi-robot multi-foraging scenario. The model's predictions
agree very closely with experimental results from sensor-based
simulations.

\end{abstract}

\section{Introduction}

In the 1980's it was considered ground-breaking for a mobile robot
to move around an unstructured environment at reasonable speeds.
In the years since, advancements in both hardware mechanisms and
software architectures and algorithms have resulted in quite
capable mobile robot systems.  Provided with this baseline
competency of individual robots, increasing attention has been
paid to the study of Multi-Robot Systems (MRS), and in particular
{\it distributed} MRS with which the remainder of this paper is
concerned.  In a distributed MRS there is no centralized control
mechanism -- instead, each robot operates independently under
local sensing and control, with coordinated system-level behavior
arising from local interactions among the robots and between the
robots and the task environment. The effective design of
coordinated MRS is restricted by the lack of formal design tools
and methodologies.  The design of single robot systems (SRS) has
greatly benefited from the formalisms provided by control theory
-- the design of MRS is in need of analogous formalisms.

For a group of robots to effectively perform a given system-level
task, the designer must address the question of which robot should
do which task and when \cite{Gerkey04}.  The process of assigning
individual robots to sub-tasks of a given system-level task is
called \textit{task allocation}, and it is a key functionality
required of any MRS.  {\it Dynamic} task allocation is a class of
task allocation in which the assignment of robots to sub-tasks is
a dynamic process and may need to be continuously adjusted in
response to changes in the task environment or group performance.
The problem of task allocation in a {\it distributed} MRS is
further compounded by the fact that task allocation must occur as
a result of a distributed process as there is no central
coordinator available to make task assignments.  This increases
the problem's complexity because, due to the local sensing of each
robot, no robot has a complete view of the world state.  Given
this incomplete and often noisy information, each robot must make
local control decisions about which actions to perform and when,
without complete knowledge of what other robots have done in the
past, are doing now, or will do in the future.

There are a number of task allocation models and philosophies.
Historically, the most popular approaches rely on intentional
coordination to achieve task allocation \cite{Parker98}.  In
those, the robots coordinate their respective actions explicitly
through deliberate communications and negotiations. Due to scaling
issues, such approaches are primarily used in MRS consisting of a
relatively small number of robots (i.e., fewer than 10). Task
allocation through intentional coordination remains the preferred
approach because it is better understood, easier to design and
implement, and more amenable to formal analysis \cite{Gerkey04}.

As the size of the MRS grows, the complexity of the design of
intentional approaches increases due to increased demands in
communication bandwidth and computational abilities of individual
robots. Furthermore, complexity introduced by increased robot
interactions makes such systems much more difficult to analyze and
design. This leads to the alternative to intentional coordination,
namely, task allocation through utilizing emergent coordination.
In systems using emergent coordination, individual robots
coordinate their actions based solely on local sensing information
and local interactions. Typically, there is very little or no
direct communication or explicit negotiations between robots. They
are, therefore, more scalable to larger numbers of robots and are
more able to take advantage of the robustness and parallelism
provided by the aggregation of large numbers of coordinated
robots.  The drawback of task allocation as achieved through
emergent coordination mechanisms is that such systems can be
difficult to design, solutions are commonly sub-optimal, and since
coordination is achieved through many simultaneous local
interactions between various subsets of robots, predictive
analysis of expected system performance is difficult.

As MRS composed of ever-larger numbers of robots become available,
the need for task allocation through emergent coordination will
increase.   To address the lack of formalisms in the design of
such MRS, in this article we present and experimentally verify a
predictive mathematical model of dynamic task allocation for MRS
using emergent coordination.  Such a formal model of task
allocation is a positive step in the direction of placing the
design of MRS on a formal footing.

The rest of the paper is organized as follows. \secref{sec:prior}
provides a summary of related work. In
\secref{sec:task-allocation} we describe a general mechanism for
task allocation in dynamic environments. This is a distributed
mechanism based on local sensing. In \secref{sec:analysis} we
present a mathematical model of the collective behavior of an MRS
using this mechanism and study its performance under a variety of
conditions. We validate the model in a multi-foraging domain.  In
\secref{sec:multi-foraging} we define the experimental task domain
of multi-foraging, robot controllers and the simulation
environment. In \secref{sec:results} we compare the predictions of
mathematical models with the results from sensor-based
simulations. We conclude the paper in \secref{sec:discussion},
with a discussion of the approach and the results.

\newcommand{\ea}{{\em et al.}}

\section{Related Work}
\label{sec:prior}

% modeling of collective behavior
Mathematical modeling and analysis of the collective behavior of
MRS is a relatively new field with approaches and methodologies
borrowed from other fields, including mathematics, physics, and
biology. Recently, a number of researchers attempted to
mathematically analyze  multi-robot systems by using
phenomenological models of the type present here. Sugawara
\ea~\cite{Sugawara97,SugSanYosAbe98} developed a simple model of
cooperative foraging in groups of communicating  and
non-co\-mmu\-ni\-cating robots. Kazadi \ea~\cite{Kazadi02} studied
the general properties of multi-robot aggregation using
phenomenological macroscopic models. Agassounon and
Martinoli~\cite{Agassounon02} presented a model of aggregation in
which the number of robots taking part in the clustering task is
based on the division of labor mechanism in ants. These models are
\emph{ad-hoc} and domain specific, and the authors give no
explanation as to how to apply such models to other domain. In
earlier works we have developed a general framework for creating
phenomenological models of collective behavior in groups of
robots~\cite{Lerman02nasa,Lerman04sab}. We applied this framework
to study collaborative stick-pulling in a group of reactive
robots~\cite{Lerman01} and foraging in robots~\cite{Lerman02}.

Most of the approaches listed above are implicitly or explicitly
based on stochastic processes theory. Another example of the
stochastic approach is the probabilistic microscopic model
developed by Martinoli and
coworkers~\cite{Martinoli99,MarIjsGam99,IMB2001} to study
collective behavior of a group of robots. Rather than compute the
exact trajectories and sensory information of individual robots,
Martinoli {\em et al.} model each robot's interactions with other
robots and the environment as a series of stochastic events, with
probabilities determined by simple geometric considerations.
Running several series of stochastic events in parallel, one for
each robot, allowed them to study the group behavior of the
multi-robot system.

\comment{ Application-level studies of adaptation and learning in
multi-robot systems have recently been carried out
~\cite{Kaelbling90,Mataric97,Martinoli02learn,Jones03icra,Dahl03icra}.
The RoboCup robot soccer domain provided a fruitful framework for
introducing learning in the context of multi-agent and multi-robot
systems. Several authors examined the use of reinforcement
learning to learn basic soccer skills, coordination
techniques~\cite{Riedmiller01} and game strategies~\cite{Stone01}.
Li {\em et al.}~\cite{Martinoli02learn} introduced learning into
collaborative stick pulling robots and showed in simulation that
learning does improve system performance by allowing robots to
specialize. No analysis of the collective behavior or performance
of the system have been attempted in any of these studies. }

So far very little work has been done on mathematical analysis of
multi-robot systems in dynamic environments. We have recently
extended~\cite{Lerman03aamas} the stochastic processes framework
developed in earlier work to robots that change their behavior
based on history of local observations of the (possibly changing)
environment~\cite{Lerman03iros}. In the current paper we develop
these ideas further, and present the exact stochastic model of the
system, in addition to the phenomenological model.

%%%%%%
Closest to ours is the work of Huberman and
Hogg~\cite{HubermanHogg88}, who mathematically studied collective
behavior of a system of adaptive agents using game dynamics as a
mechanism for adaptation.  In game dynamical systems, winning
strategies are rewarded, and agents use the best performing
strategies to decide their next move. Although their adaptation
mechanism is different from our dynamic task allocation mechanism,
their analytic approach is similar to ours, in that it is based on
the theory of stochastic processes. Others have mathematically
studied collective behavior of systems composed of large numbers
of concurrent learners~\cite{Wolpert99,Sato03}. These are
microscopic models, which only allow one to study collective
behavior of relatively small systems of a few robots. We are
interested in macroscopic approaches that enable us to directly
study collective behavior in large systems. Our work differs from
earlier ones in another important way:  we systematically compare
theoretical predictions of mathematical models with results of
experiments carried out in a sensor-based simulator.

\section{Dynamic Task Allocation Mechanism\label{sec:task-allocation}}

The dynamic task allocation scenario we study considers a world
populated with tasks of $T$ different types and robots that are
equally capable of performing each task but can only be assigned
to one type at any given time. For example, the tasks could be
targets of different priority that have to be tracked, different
types of explosives that need to be located, etc. Additionally, a
robot cannot be idle --- each robot is always performing a task at
any given time. We introduce the notion of a robot state as a
shorthand for the type of task the robot is assigned to service. A
robot may switch its state according to its control policy when it
determines it is appropriate to do so. However, needlessly
switching tasks is to be avoided, since in physical robot systems,
this can involve complex physical movement that requires time to
perform.

The purpose of task allocation is to assign robots to tasks in a
way that will enhance the performance of the system, which
typically means reducing the overall execution time. Thus, if all
tasks take an equal amount of time to complete, in the best
allocation, the fraction of robots in state $i$ will be equal to
the fraction of tasks of type $i$. In general, however, the
desired allocation could take other forms ---  for example, it
could be related to the relative reward or cost of completing each
task type --- without change to our approach. In the dynamic task
allocation scenario, the number of tasks and the number of
available robots are allowed to change over time, for example, by
adding new tasks, deploying new robots, or removing malfunctioning
robots.

The challenge faced by the designer is to devise a mechanism that
will lead to a desired task allocation in a distributed MRS even
as the environment changes. The challenge is made even more
difficult by the fact that robots have limited sensing
capabilities, do not directly communicate with other robots, and
therefore, cannot acquire global information about the state of
the world, the initial or current number of tasks (total or by
type), or the initial or current number of robots (total or by
assigned type). Instead, robots can sample the world (assumed to
be finite) --- for example, by moving around and making local
observations of the environment. We assume that robots are able to
observe tasks and discriminate their types. They may also be able
to observe and discriminate the task states of other robots.

One way to give the robot an ability to respond to environmental
changes (including actions of other robots) is to give a robot an
internal state where it can store its knowledge of the environment
as captured by its observations~\cite{Jones03iros,Lerman03aamas}.
The observations are stored in a rolling history window of finite
length, with new observations replacing the oldest ones. The robot
consults these observations periodically and updates its task
state according to some transition function specified by the
designer. In an earlier work we
showed~\cite{Jones03iros,Lerman03iros} that this simple dynamic
task allocation mechanism leads to the desired task allocation in
a multi-foraging scenario.

In the following sections we present a mathematical model of
dynamic task allocation and study the role that transition
function and the number of observations (history length) play in
the performance of a multi-foraging MRS. In
\secref{sec:pucksonly}, we present a model of a simple scenario in
which robots base their decisions to change state solely on
observations of tasks in the environment. We study the simplest
form of the transition function, in which the probability to
change state to some type is proportional to the fraction of
existing tasks of that type. In \secref{sec:results1} we compare
theoretical predictions with no adjustable parameters to
experimental data and find excellent agreement. In
\secref{sec:phenomenological} we examine the more complex scenario
where the robots base their decisions to change task state on the
observations of both existing task types and task states of other
robots. In \secref{sec:results2} we study the consequences of the
choice of the transition function and history length on the system
behavior and find good agreement with the experimental data.

\section{Analysis of Dynamic Task Allocation}
\label{sec:analysis}

As proposed in the previous section,  a robot may be able to adapt
to a changing environment in the absence of complete global
knowledge if it is  able to make and remember local observations
of the environment. In the treatment below we assume that there
are two types of tasks
---  arbitrarily referred to as $Red$ and $Green$. This simplification is for pedagogical reason only; the model
can be extended to a greater number of task types.

During a sufficiently short time interval, each robot can be
considered to belong to the $Green$ or $Red$ task state. This is a
very high level, coarse-grained description.  In reality, each
state is composed of several robot actions and behaviors, for
example, searching for new tasks, detecting and executing them,
avoiding obstacles, {\em etc}. However, since we want the model to
capture how the fraction of robots in each task state evolves in
time, it is a sufficient level of abstraction to consider only
these two states. If we find that additional levels of detail are
required to explain system behavior, we can elaborate the model by
breaking each of the high level states into its underlying
components.

\subsection{Observations of Tasks Only}
\label{sec:pucksonly}

In this section we study dynamic task allocation mechanism in
which robots make decisions to switch task states based solely on
observations of available tasks. Let $m_r$ and $m_g$ be  the
numbers of the observed $Red$ and $Green$ tasks, respectively, in
a robot's memory or history window.  The robot chooses to change
its state, or the type of task it is assigned to execute, with
probabilities given by transition functions $f_{g \rightarrow
r}(m_r,m_g)$ (probability of switching to $Red$ from $Green$) and
$f_{r \rightarrow g}(m_r,m_g)$ (probability of switching to
$Green$ from $Red$). We would like to define transition rules so
that the fraction of time the robot spends in the $Red$ ($Green$)
state be equal to the fraction of $Red$ ($Green$) tasks. This will
assure that on average the number of $Red$  and $Green$ robots
reflect the desired task distribution. Clearly, if the robots have
global knowledge about the numbers of $Red$ and $Green$ tasks
$M_r$ and $M_g$, then each robot could choose each state with
probability equal to the fraction of the tasks of corresponding
type. Such global knowledge is not available; hence, we want to
investigate how incomplete knowledge of the environment (through
local observations), as well as the dynamically changing
environment (e.g., changing ratio of $Red$ and $Green$ tasks),
affects task allocation.

\subsubsection{Modelling Robot Observations}

As explained above, the transition rate between task execution
states depends on robot's observations stored in its history. In
our model we assume that a robot makes an observation of a task
with a time period $\tau$. For simplicity, by an observation we
mean here detecting a task, such as a target to be monitored, mine
to be cleared or an object to be gathered. Therefore, observation
history of length $h$ comprises of the number of $Red$ and $Green$
tasks a robot has observed during a time interval ${h}\tau$. We
assume that $\tau$ has unit length and drop it. The process of
observing a task is given by a Poisson distribution with rate
$\lambda = \alpha M^0$, where $\alpha$ is a constant
characterizing the physical parameters of the robot such as its
speed, view angles, etc., and $M^0$ is the number of tasks in the
environment. This simplification is based on the idea that robot's
interactions with other robots and the environment are independent
of the robot's actual trajectory and are  governed by
probabilities determined by simple geometric considerations. This
simplification has been shown to produce remarkably good
agreements with experiments~\cite{MarIjsGam99,IMB2001}.

Let $M_r(t)$ and $M_g(t)$ be the number of $Red$ and $Green$ tasks
respectively (can be time dependent), and let
$M(t)=M_r(t) + M_g(t)$ be the total number of tasks. The
probability that in the time interval $[t-{h}, t]$ the robot has
observed exactly $m_r$ and $m_g$ tasks is the product of two
Poisson distributions:
\begin{equation}
\label{eq:poisson} P(m_r,m_g) = \frac{\lambda_r^{m_r}
\lambda_g^{m^g}}{m_r! m_g!}e^{- \lambda_r-\lambda_g}
\end{equation}
where $\lambda_i$~, $i=r,g$, are the means of the respective
distributions. If the task distribution does not change in time,
$\lambda_i = \alpha M_i {h}$. For time dependent task
distributions, $\lambda_i = \alpha \int_{t-{h}}^tdt'M_i(t')$.

\subsubsection{Individual Dynamics: The Stochastic Master Equation}

Let us consider a single robot that has to decide between
executing  $Red$ and $Green$ tasks in a closed arena and makes a
transition to $Red$ and $Green$ states according to its
observations. Let $p_r(t)$ be the probability that a robot is in
the $Red$ state at time $t$. The equation governing its evolution
is
\begin{equation}
\label{eq:inddyn} \frac{dp_r}{dt} = \varepsilon (1-p_r) f_{g
 \rightarrow r} - \varepsilon p_r f_{r \rightarrow
g}
\end{equation}
where $\varepsilon$ is the rate at which the robot makes decisions
to switch its state, and $f_{g  \rightarrow r}$ and $f_{r
\rightarrow g}$ are the corresponding transitions probabilities
between the states. As explained above, these probabilities depend
on the robot's history --- the number of tasks of either type it
has observed during the time interval ${h}$ preceding the
transition. If the robots have global knowledge about the numbers
of $Red$ and $Green$ tasks $M_r$ and $M_g$, one could choose the
transition probabilities as the fraction of tasks of corresponding
type, $f_{g \rightarrow r}\propto M_r/(M_r+M_g)$ and $f_{r
\rightarrow g}\propto M_g/(M_r+M_g)$. In the case when the global
information is not available, it is natural to use similar
transition probabilities using robots' local estimates:
\begin{eqnarray}
\label{eq:rates} f_{g \rightarrow r}(m_r,m_g) =
\frac{m_r}{m_r+m_g}\equiv
\gamma_r(m_r,m_g) \\
\label{eq:ratesg} f_{r \rightarrow g}(m_r,m_g) =
\frac{m_g}{m_r+m_g}\equiv
 \gamma_g(m_r,m_g)
\end{eqnarray}
Note that $\gamma_r(m_r,m_g) + \gamma_g(m_r,m_g) =1$ whenever
$m_r+m_g>0$, \eg, whenever there is at least one observation in
the history window. In the case when there are no observations in
history, $m_r=m_g=0$, robots will choose either state with
probability $1/2$ as it follows from taking the appropriate limits
in Equations \ref{eq:rates} and \ref{eq:ratesg}. Hence, we
supplement \eqref{eq:rates} with $f_{g \rightarrow r}(0,0)=f_{r
\rightarrow g}(0,0)=0$ (and similarly for \eqref{eq:ratesg}) to
assure that robots do not change their state when the history
window does not contain any observations.

\eqref{eq:inddyn}, together with the transition rates shown in
Equations \ref{eq:rates}--\ref{eq:ratesg}, determines the
evolution of the probability density of a robot's state. It is a
stochastic equation since the coefficients (transition rates)
depend on random variables $m_r$ and $m_g$. Moreover, since the
robot's history changes gradually, the values of the coefficients
at different times are correlated, hence making the exact
treatment very difficult. We propose, instead, to study the
problem within the $annealed$ approximation: we neglect
time--correlation between robot's histories at different times,
assuming instead that at any time the real history $\{m_r,m_g\}$
can be replaced by a random one drawn from the Poisson
distribution \eqref{eq:poisson}. Next, we average
\eqref{eq:inddyn} over all histories to obtain
\begin{equation}
\label{eq:inddyn2} \frac{dp_r}{dt} = \varepsilon
\overline{\gamma}_r (1-n_r)  - \varepsilon \overline{\gamma}_g n_r
\end{equation}
Here $\overline{\gamma}_r$ and $\overline{\gamma}_g$ are given by
\begin{equation}
\label{eq:avg_gamma} \overline{\gamma}_r =
\sum_{r,g}P(r,g)\frac{r}{r+g},  \overline{\gamma}_g =
\sum_{r,g}P(r,g)\frac{g}{r+g}
\end{equation}
where $P(m_r,m_g)$ is the Poisson distribution \eqref{eq:poisson}
and the summation excludes the term $r=g=0$. Note that if the
distribution of tasks changes in time, then
$\overline{\gamma}_{r,g}$ are time-dependent,
$\overline{\gamma}=\overline{\gamma}_{r,g}(t)$.

To proceed further, we need to evaluate the summations in
\eqref{eq:avg_gamma}. Let us  define an auxiliary function
\begin{eqnarray}
\label{eq:aux} F(x) =
\sum_{m_r=0}^{\infty}\sum_{m_g=0}^{\infty}{x^{m_r+m_g} \frac
{\lambda_r^{m_r} \lambda_g^{m_g}} {m_r! m_g!}
e^{-\lambda_r}e^{-\lambda_g}\frac {m_r} {m_r+m_g}}
\end{eqnarray}
It is easy to check that $\overline{\gamma}_{r,g}$ are given by
\begin{eqnarray}
\label{eq:aux1} \overline{\gamma}_r &=& F(1) - \frac{1}{2}P(0,0)
= F(1) - \frac{1}{2} e^{\alpha {h} M_0} \nonumber \\
\overline{\gamma}_g &=& 1-F(1) -\frac{1}{2}e^{\alpha {h} M_0}
\end{eqnarray}
Differentiating \eqref{eq:aux} with respect to $x$ yields
\begin{equation}
\label{eq:aux2} \frac{dF}{dx} =
\sum_{m_r=1}^{\infty}\sum_{m_g=0}^{\infty}x^{m_r+m_g-1}\frac{\lambda_r^{m_r}
\lambda_g^{m_g}}{m_r! m_g!}e^{- \lambda_r}e^{-\lambda_g}m_r
\end{equation}
Note that the summation over $m_r$ starts from $m_r=1$. Clearly,
 the sums over $m_r$ and $m_g$ are de--coupled thanks to the
cancellation of the denominator $(m_r+m_g)$:
\begin{equation}
\label{eq:aux3} \frac{dF}{dx} = \biggl( e^{-
\lambda_r}\sum_{m_r=1}^{\infty}x^{m_r-1}\frac{\lambda_r^{m_r}
}{m_r!}m_r \biggr )\biggl( e^{-
\lambda_g}\sum_{m_g=0}^{\infty}\frac{(x \lambda_g)^{m_g} }{m_g!}
\biggr )
\end{equation}
The resulting sums are evaluated easily (as the Taylor expansion
of corresponding exponential functions), and the results is
\begin{equation}
\label{eq:diff} \frac{dF}{dx} = \lambda_r e^{- \lambda_0(1-x)}
\end{equation}
where $\lambda_0 = \lambda_r+\lambda_g$. After elementary
integration of \eqref{eq:diff} (subject to the condition $F(0) =
1/2$),  we obtain using \eqref{eq:aux2} and the expressions for
$\lambda_r$, $\lambda_0$:
\begin{equation}
\label{eq:gamma} \overline{\gamma}_{r,g}(t) = \frac{1-e^{\alpha
{h} M_0}}{{h} }\int_{t-{h}}^t dt^{\prime}\mu_{r,g}(t^{\prime})
\end{equation}
Here $\mu_{r,g}(t)=M_{r,g}(t)/M_0$ are the fraction of $Red$ and
$Green$ tasks respectively.

Let us first consider the case when the task distribution does not
change with time, \ie, $\mu_r(t)=\mu_0$. Then we have
\begin{equation}
\label{eq:gamma-constant} \overline{\gamma}_{r,g}(t)
=(1-e^{-\alpha {h} M_0})\mu_{r,g}^0
\end{equation}
The solution of \eqref{eq:inddyn2} subject to the initial
condition $p_r(t=0)=p_0$ is readily obtained:
\begin{equation}
\label{eq:solution} p_r(t) = \mu_r^0 + \biggl ( p_0 -
\frac{\overline{\gamma}_r}{\overline{\gamma}_r +
\overline{\gamma}_g}\biggr )e^{ - \varepsilon (\overline{\gamma}_r
+ \overline{\gamma}_g)t}
\end{equation}
One can see that the probability distribution approaches  the
desired steady state value $p_r^s = \mu_r^0$ exponentially. Also,
the coefficient of the exponent depends on the density of tasks
and the length of the history window. Indeed, it is easy to check
that $\overline{\gamma}_r + \overline{\gamma}_g = 1-e^{-\alpha
{h} M_0}$. Hence,   for large enough $M_0$  and ${h}$, $\alpha
{h} M_0 \gg 1$, the convergence rate is determined solely by
$\varepsilon$. For a small task density or short history
length, on the other hand, the convergence rate is proportional to
the number of tasks, $ \varepsilon ( 1-e^{-\alpha {h} M_0})\sim
\varepsilon \alpha {h} M_0$. Note that this is a direct
consequence of the rule that robots do not change their state
whenever there are no observation in the history window.

Now let us consider the case where the task distribution changes
suddenly at time $t_0$, $\mu_r(t)= \mu_r^0 + \Delta \mu
\theta(t-t_0)$, where $\theta(t-t_0)$ is the step function. For
simplicity, let us assume that $\alpha {h} M_0 \gg 1$ so that the
exponential term in \eqref{eq:gamma} can be neglected,
\begin{equation}
\label{eq:gamma-timedep} \overline{\gamma}_{r,g}(t) =
\frac{1}{{h} }\int_{t-{h}}^t dt^{\prime}\mu_{r,g}(t^{\prime}),
\overline{\gamma}_{r}(t) + \overline{\gamma}_{g} =1
\end{equation}

Replacing \eqref{eq:gamma-timedep} into \eqref{eq:inddyn2}, and
solving the resulting  differential equation yields
\begin{eqnarray}
\label{eq:jump} p_r(t)& =& \mu_r^0 + \frac{\Delta\mu}{{h}}t
-\frac{\Delta\mu}{\varepsilon {h}}(1-e^{-\varepsilon
t}),~~~~~~~~~~t\leq {h} \nonumber \\
p_r(t)& =& \mu_r^0 + \Delta\mu -\frac{\Delta\mu}{\varepsilon
{h}}(e^{-\varepsilon (t-{h})} -e^{-\varepsilon t} ),~~~t>{h}
\,.
\end{eqnarray}
\eqref{eq:jump} describes how the robot distribution converges to the
new steady state value after the change in task distribution.
Clearly, the convergence properties of the solutions depend on
${h}$ and $\varepsilon$. It is easy to see that in the limiting
case $\varepsilon {h} \gg 1$ the new steady state is attained
after time ${h}$, $|p_r({h})-(\mu_0 + \Delta \mu)| \sim \Delta
\mu/(\varepsilon {h})\ll 1$, so the convergence time is
$t_{conv}\sim {h}$. In the other limiting case $\varepsilon {h}
\ll 1$, on the other hand, the situation is different. A simple
analysis of \eqref{eq:jump} for $t>{h}$ yields $|p_r(t)-(\mu_0
+ \Delta \mu)| \sim \Delta \mu e^{-\varepsilon t}$ so the
convergence is exponential with characteristic time $t_{conv} \sim
1/\varepsilon$.

\subsubsection{Collective Behavior}
In order to make predictions about the behavior of an MRS using a
dynamic task allocation mechanism, we need to develop a
mathematical model of the collective behavior of the system. In
the previous section we derived a model of how an individual
robot's behavior changes in time. In this section we extend it to
model the behavior of a MRS. In particular, we study the
collective behavior of a homogenous system consisting of $N$
robots with identical controllers. Mathematically, the MRS is
described by a probability density function that includes the
states of all $N$ robots. However, in most cases we are interested
in studying the evolution of global, or average, quantities, such
as the average number of robots in the $Red$ state, rather than
the exact probability density function. This applies when
comparing theoretical predictions with results of experiments,
which are usually quoted as an average over many experiments.
Since the robots in either state are independent of each other,
$p_r(t)$, is now the fraction of robots in the $Red$ state, and
consequently $Np_r(t)$ is the average number of robots in that
state. The results of the previous section, namely solutions for
$p_r(t)$ for constant task distribution (\eqref{eq:solution}) and
for changing task distribution (\eqref{eq:jump}), can be used to
study the average collective behavior. \secref{sec:results1}
presents results of analysis of the mathematical model.

\subsubsection{Stochastic Effects}
\label{sec:stochastic1}
In some cases it is useful to know the probability distribution of
robot task states over the entire MRS. This probability function describes
the exact collective behavior from which one could derive the
average behavior as well as the fluctuations around the average.
Knowing the strength of fluctuations is necessary for assessing how
the probabilistic nature of robot's observations and actions affects the global
properties of the system. Below we consider the problem of finding
the probability distribution of the collective state of the
system.

Let $P_n(t)$ be the probability that there are exactly $n$ robots
in the $Red$ state at time $t$. For a sufficiently short time
interval $\Delta t$ we can write~\cite{Lerman03iros}
\begin{equation}
\label{eq:master1} P_n(t+\Delta t) = \sum_{n^{\prime}}
W_{n^{\prime} n}(t;\Delta t)P_{n^{\prime}}(t) -\sum_{n^{\prime}}
W_{n n^{\prime}}(t;\Delta t)P_{n}(t)
\end{equation}
where $W_{i j}(t;\Delta t)$ is the transition probability between
the states $i$ and $j$ during the time interval $(t, t+\Delta t)$.
In our MRS, this transitions correspond to robots changing their
state from $Red$ to $Green$ or vice versa. Since the probability
that more than one robot will have a transition during a time
interval $\Delta t$ is $O(\Delta t)$, then, in the limit $\Delta t
\rightarrow 0$ only transitions between neighboring states are
allowed in \eqref{eq:master1}, $n \rightarrow n \pm 1$. Hence, we
obtain

\begin{equation}
\label{eq:master2} \frac{dP_n}{dt} = r_{n+1}P_{n+1}(t) +
g_{n-1}P_{n-1}(t) - (r_n+g_n)P_n(t) \,.
\end{equation}
Here $r_k$ is the probability density of having one of the $k$
$Red$ robots change its state to $Green$, and $g_k$ is the
probability density of having one of the $N-k$ $Green$ robots
change its state to $Red$. Let us assume again that $\alpha {h}
M_0 \gg 1$ so that $\overline{\gamma}_g = 1-\overline{\gamma}_r$.
Then one has
\begin{equation}
r_k = k  (1 - \overline{\gamma}_r ) \ , \ g_k = (N-k)
\overline{\gamma}_r
\end{equation}
with $r_0=g_{-1}=0$, $r_{N+1}=g_N = 0$. $\overline{\gamma}_r$ is
history-averaged transition rate to $Red$ states.

The steady state solution of \eqref{eq:master2} is given by
\cite{VanKampen}
\begin{equation}
P_n^s = \frac{g_{n-1}g_{n-2}...g_1g_0}{r_nr_{n-1}...r_2r_1}P_0^s
\end{equation}
where $P_0^s$ is determined by the normalization:
\begin{equation}
P_0^s = \left[ 1+ \sum_{n=1}^{N}
\frac{g_{n-1}g_{n-2}...g_1g_0}{r_nr_{n-1}...r_2r_1} \right]^{-1}
\end{equation}
Using the expression for $\overline{\gamma}$,  after
some algebra we obtain
\begin{equation}
\label{eq:Pn} P_n^s = \frac{N!}{(N-n)!n!}
\overline{\gamma}_r^n(1-\overline{\gamma}_r)^{N-n}
\end{equation}
e.g., the steady state is a binomial distribution with parameter
$\overline{\gamma}$. Note again that this is a direct consequence
of the independence of the robots' dynamics. Indeed, since the
robots act independently, in the steady state each robot has the
same probability of being in either state. Moreover, using this
argument it becomes clear that the time-dependent probability
distribution $P_n(t)$ is given by \eqref{eq:Pn} with
$\overline{\gamma}$ replaced by  $p_r(t)$, \eqref{eq:solution}.

\subsection{Observations of Tasks and Robots}
\label{sec:phenomenological} In this section we study the more
complex dynamic task allocation mechanism in which robots make
decisions to change their state based on the observations of not
only available tasks but also on the observed task states of other
robots. Specifically, each robot now records the numbers and types
of task as well as the numbers and task types of robots it has
encountered. Again, we let $m_r$ and $m_g$ be the number of tasks
of $Red$ and $Green$ type, and $n_r$ and $n_g$ be the number of
robots in $Red$ and $Green$ task state in a robot's history
window. The probabilities for changing a robot's state are again
given by transition functions that now depend on the fractions of
observed tasks and robots of each type: $\hat{m}_r=m_r/(m_r+m_g)$,
$\hat{m}_g=m_g/(m_r+m_g)$, $\hat{n}_r=n_r/(n_r+n_g)$, and
$\hat{n}_g=n_g/(n_r+n_g)$. In our previous
work~\cite{Lerman03iros} we showed that in order to achieve the
desired long term behavior for task allocation (\ie, in the steady
state the average fraction of $Red$ and $Green$ robots is equal to
the fraction of $Red$ and $Green$ tasks respectively), the
transition rates must have the following functional form:
\begin{eqnarray}
\label{eq:fR}
f_{g \rightarrow r} (\hat{m}_r,\hat{n}_r) &=&\hat{m}_rg(\hat{m}_r-\hat{n}_r),\\
\label{eq:fG} f_{r \rightarrow g} (\hat{m}_r,\hat{n}_r)
&=&\hat{m}_g g(\hat{m}_g-\hat{n}_g) \equiv (1-\hat{m}_r)
g(-\hat{m}_r+\hat{n}_r) .
\end{eqnarray}
Here $g(z)$ is a continuous, monotonically increasing function of
its argument defined on an interval $[-1,1]$. In this paper we
consider the following forms for $g(z)$:
\begin{itemize}
\item {\em Power:}
$g(z)=100^{z}/100$
\item  {\em Stepwise linear:} $g(z)=z \Theta(z)$.\footnote{The step function $\Theta$ is defined as
$\Theta(z)=1$ if $z \geq 0$; otherwise, it is $0$. The step
function guarantees that no transitions to $Red$ state occur when
$ {m}_r < {n}_r$.}
\end{itemize}

To analyze this task allocation model, let us again consider a
single robot that searches for tasks to perform and makes a
transition to $Red$ and $Green$ states according to transition
functions defined above. Let $p_r(t)$ be the probability that the
robot is in the $Red$ state at time $t$, with \eqref{eq:inddyn}
governing its time evolution. Note that $p_r(t)$ is also the
average fraction of $Red$ robots, $p_r(t) = N_r(t)/N$.
\comment{
The equation governing the evolution of $p_r(t)$ reads
%\begin{equation}
%\label{eq:inddyn3} \frac{dp_r}{dt} = \varepsilon (1-p_r) f_{g
%\rightarrow r} - \varepsilon p_r f_{r \rightarrow g}
%\end{equation}
Here, again, $\varepsilon$ is the rate at which the robot makes
decisions to switch its state}

As in the previous case, the next step of the analysis is
averaging over the the robot's histories, \ie, $\hat{m}_r$ and
$\hat{n}_r$. Note that a robot's observations of available tasks
can still be modeled by a Poisson distribution similar to
\eqref{eq:poisson}. However, since the number of robots of each
task state changes stochastically in time, the statistics of $n_r$
and $n_g$
 should be modeled as a doubly stochastic
Poisson process (also called Cox process) with stochastic rates.
This would complicate the calculation of the average over $\hat{n}_r =
n_r/(n_r+n_g)$ and require mathematical details that go well
beyond the scope of this paper. Fortunately, as we demonstrated in
the previous section, if a robot's observation window contains
many readings, then the estimated fraction of task types is
exponentially close to the average of the Poisson distribution.
This suggests that for sufficiently high densities of tasks and
robots we can neglect the stochastic effects of modeling
observations for the purpose of our analysis, and replace the
robot's observation by their average (expected) values. In other
words, we use the following approximation:
\begin{eqnarray}
\label{eq:nr2}
\hat{n}_r &\approx& \frac{1} {h} \int_{t-h}^{t}{p_r(t')}dt'\\
\label{eq:mr2} \hat{m}_r &\approx& \frac{1} {h}
\int_{t-h}^{t}{\mu_r(t')}dt' .
\end{eqnarray}

The  Equations~\ref{eq:inddyn}, ~\ref{eq:nr2}, and~\ref{eq:mr2}
are a system of integro--differential equations that uniquely
determine the dynamics of $p_r(t)$. In the most general case it is
not possible to obtain solutions by analytical means, hence one
has to solve the system numerically. However, if the task density
does not change in time, we can still perform steady state
analysis. Steady state analysis looks for long-term solutions that
do not change in time, \ie, $d p_r/dt=0$. Let $\mu_r^0$ be the
density of $Red$ tasks, and $p_0=p_r(t \rightarrow \infty)$ be the
steady state value, so that $\hat{m}_r = \mu_r^0$, $\hat{n}_r =
p_r^0$. Then, by setting left hand side of \eqref{eq:inddyn} to
zero, we get

\begin{equation}
\label{eq:steady} (1-p_0) \mu_r^0 g(\mu_r^0-p_0) = p_0 (1-\mu_r^0)
g(-\mu_r^0+p_0)
\end{equation}

Note that $p_0=\mu_r^0$ is a solution to \eqref{eq:steady} so that
in the steady state the fraction of $Red$ robots equals the fraction
of red tasks as desired. To show that this is the only solution,
we note that for a fixed $\mu_r^0$ the right- and left-hand sides
of the equation are monotonically increasing and decreasing
functions of $p_0$ respectively, due to the monotonicity of $g(z)$.
Consequently, the two curves can meet only once and that proves the
uniqueness of the solution.

\subsubsection{Phenomenological Model}
\label{sec:stochastic}

Exact stochastic models of task allocation can quickly become
analytically intractable, as we saw above. Instead of exact
models, it is often more convenient to work with the so-called
Rate Equations model. These equations can be derived from the
exact stochastic model by appropriately averaging
it~\cite{Lerman03iros}; however, they are often (see, for example,
population dynamics~\cite{Haberman}) phenomenological, or \emph{ad
hoc}, in nature --- constructed by taking into account the
system's salient processes. This approach makes a number of
simplifying assumptions: namely, that the system is uniform and
dilute (not too dense), that actions of individual entities are
independent of one another, that parameters can be represented by
their mean values and that system behavior can be described by its
average value. Despite these simplifications, resulting models
have been shown to correctly describe dynamics of collective
behavior of robotic systems~\cite{Lerman04sab}. Phenomenological
models are useful for answering many important questions about the
performance of a MRS, such as, does the steady state exist, how
long does it take to reach it, and so on. Below we present a
phenomenological model of dynamic task allocation.

Individual robots are making their decisions to change task state
probabilistically and independently of one another. A robot will
change state from $Green$ to $Red$ with probability $f_{g
\rightarrow r}$ and with probability $1-f_{g \rightarrow r}$ it
will remain in the $Green$ state. We can succinctly write $\Delta
N_{g \rightarrow r}$  and $\Delta N_{r \rightarrow g}$, the number
of robots that switch from $Green$ to $Red$ and \emph{vice versa}
during a sufficiently small time interval $\Delta t$, as
\begin{eqnarray}
\Delta N_{g \rightarrow r}& =&\sum_{i=1}^{N_g}{x_i \big(f_{g
\rightarrow r} \delta(x_i-1)+(1- f_{g \rightarrow r})
\delta(x_i)\big)}
\nonumber\\
\Delta N_{r \rightarrow g}&=&\sum_{i=1}^{N_r}{(1-x_i) \big(f_{r
\rightarrow g} \delta(x_i)+(1-f_{r \rightarrow g})
\delta(x_i-1)\big)}\,. \nonumber
\end{eqnarray}
Here we introduced a state variable $x_i$, such that $x_i=1$ when a robot is in the $Green$ state,
and $x_i=0$ when a robot is in the $Red$ state. $\delta(x)$ is Kronecker delta, defined as $\delta(x)=1$ when $x=0$ and $\delta(x)=0$ otherwise.  Therefore, $\Delta N_{g \rightarrow r}$
is a random variable from a binomial distribution specified by a mean $\mu=f_{g \rightarrow r} N_g$ and variance
$\sigma^2=f_{g \rightarrow r} (1-f_{g \rightarrow r}) N_g$.
Similarly, the distribution of the random variable $\Delta N_{r
\rightarrow g}$ is specified by mean $\mu=f_{r \rightarrow g} N_r$
and variance $\sigma^2=f_{r \rightarrow g} (1-f_{r \rightarrow g})
N_r$.

During a time interval $\Delta t$ the total number of robots in
$Red$ and $Green$ task states will change as individual
robots make decisions to change states. The following finite
difference equation specifies how the number of $Red$ will change on average:
\begin{equation}
\label{eq:stochastic} N_r(t+\Delta t)  = N_r(t) + \varepsilon
\Delta N_{g \rightarrow r} \Delta t - \varepsilon  \Delta N_{r
\rightarrow g} \Delta t \nonumber
\end{equation}
Rearranging the equation and taking the continuous time limit
($\Delta t \rightarrow 0$) yields a differential Rate Equation
that describes time evolution of the number of $Red$ robots. By
taking the means of $\Delta N$'s as their values, we recover
\eqref{eq:inddyn}.

Keeping $\Delta N$'s as random variables allows us to study the
effect the probabilistic nature of the robots' decisions have on
the collective behavior.\footnote{Note that we do not model here
the effect of observation noise due to uncertainty in sensor
readings and fluctuations in the distribution of tasks.} We solve
\eqref{eq:stochastic} by iterating it in time and drawing $\Delta
N$'s at random from their respective distributions. The solutions
are subject to the initial condition $N_r(t \leq 0)=N$ and specify
the dynamics of task allocation in robots.

%%%%%%%%%%%%%%%%%%%%%%%

Functions $f_{g \rightarrow r}$ and $f_{r \rightarrow g}$ are
calculated using
estimates of the densities of $Red$ tasks ($m_r$) and robots in $Red$ state ($n_r$)
from the observed counts stored in the robot's history window.

% collective transition rates
Transition rates  $f_{g \rightarrow r}$ and $f_{r \rightarrow g}$ in the model are mean values, averaged over all histories and all robots. In order to compute them, we need to aggregate observations of all robots. Suppose each robot has a history  window of length $h$.
For a particular robot $i$, the values in the most recent observational slot
are $N^0_{i,r}$, $N^0_{i,g}$, $M^0_{i,r}$ and
$M^0_{i,g}$, the observed numbers of $Red$ and $Green$ robots
and tasks respectively at time $t$. In the next latest slot, the
values are $N^1_{i,r}$, $N^1_{i,g}$, $M^1_{i,r}$ and
$M^1_{i,g}$, the observed numbers at time $t-\Delta$, and so
on. Each robot estimates the densities of $Red$ robots and tasks using the following calculation:
\begin{eqnarray}
{n}_{i,r} & = & \frac{1}{h}\sum^{h-1}_{j=0}{\frac{N^j_{i,r}}{N^j_{i,r}+N^j_{i,g}}}= \frac{1}{h}\sum^{h-1}_{j=0}{n^j_{i,r}} \\
{m}_{i,r} & = & \frac {1}{h}
\sum^{h-1}_{j=0}{\frac{M^j_{i,r}}{M^j_{i,r}+M^j_{i,g}}}=
\frac{1}{h}\sum^{h-1}_{j=0}{m^j_{i,r}}.
\end{eqnarray}

When observations of all robots are taken into account, the
mean of the observed densities of $Red$ robots at time $t$ --- $\frac{1} {N} \sum^N_{i=1}{n^0_{i,r}} $  --- will fluctuate due to observation noise, but on average it will be proportional to $N_r(t)/N$, which is the actual density of $Red$
robots at time $t$. The proportionality factor is related to
physical robot parameters, such as speed and observation area (see \secref{sec:results1}).
Likewise, the average of the observed densities at time  $t-j \Delta$ is $\frac{1} {N}
\sum^N_{i=1}{n^j_{i,r}} \propto N_r(t-j\Delta)/N$, the
 density of robots at time $t-j\Delta$.  Thus, the aggregate
estimates of the fractions of $Red$ robots and tasks are:
\begin{eqnarray}
\label{eq:2} \hat{n}_r & = & \frac{1}{N}
\sum^{N}_{i=1}{{n}_{i,r}}=
\frac{1}{Nh}\sum^{h-1}_{j=0}{N_r(t-j\Delta)} \\
\label{eq:2b}\hat{m}_r & = & \frac{1}{N}
\sum^{N}_{i=1}{{m}_{i,r}}=
\frac{1}{Mh}\sum^{h-1}_{j=0}{M_r(t-j\Delta)}
\end{eqnarray}
Robots are making their decisions asynchronously, {\em i.e.}, at
slightly different times. Therefore, the last terms in the above
equations are best expressed in continuous form: {\em e.g.}, $1/Nh
\int^0_{h}N_r(t-\tau)d\tau$ (see \eqref{eq:nr2} and
\eqref{eq:mr2}).

Estimates \eqref{eq:2} and \ref{eq:2b} can be plugged into
\eqref{eq:fR} and \eqref{eq:fG} to compute the values of
transition probabilities for any choice of the transition function
(power or linear). Once we know $f_{r \rightarrow g}$ and $f_{g
\rightarrow r}$, we can solve \eqref{eq:stochastic} to study the
dynamics of task allocation in robots. Note that
\eqref{eq:stochastic} is now a time-delay finite difference
equation, and solutions will show typical oscillations.

We solve the models presented in this section and validate their predictions in context of the multi-foraging task described next.

\section{Multi-Robot Multi-Foraging Task\label{sec:multi-foraging}}

In this section we describe the multi-foraging task domain in
which we experimentally tested our dynamic task allocation
mechanism, including the simulation environment used and robot
sensing and control characteristics. In \secref{sec:results1} we
use this application to validate the models presented above, solve
them and compare their solutions to the results of embodied
simulations.

\subsection{Task Description \label{sec:task_description}}

The traditional foraging task is defined by having an individual robot or group of robots collect a set of objects from an environment and either consume on the spot or return them to a common location \cite{Goldberg02}.  Multi-foraging, a variation on traditional foraging, is defined in \cite{Balch99} and consists of an arena populated by multiple types of objects to be concurrently collected.

In our multi-foraging domain, there are two types of objects (e.g., pucks) randomly dispersed throughout the arena: Puck\(_{Red}\) and Puck\(_{Green}\) pucks that are distinguishable by their color.  Each robot is equally capable of foraging both puck types, but can only be allocated to foraging for one type at any given time.  Additionally, all robots are engaged in foraging at all times; a robot cannot be idle.  A robot may switch the puck type for which it is foraging according to its control policy, when it determines it is appropriate to do so. This is an instantiation of the general task allocation problem described earlier in this paper, with puck colors representing different task types.

In the multi-foraging task, the robots move in an enclosed arena
and pick up encountered pucks.  When a robot picks up a puck, the
puck is consumed (i.e., it is immediately removed from the
environment, not transported to another region) and the robot
carries on foraging for other pucks.  Immediately after a puck is
consumed, another puck of the same type is placed in the arena at
a random location.  This is done so as to maintain a constant puck
density in the arena throughout the course of an experiment.  In
some situations, the density of pucks can impact the accuracy or
speed of convergence to the desired task allocation.  This is an
important consideration in dynamic task allocation mechanisms for
many domains; however, in this work we want to limit the number of
experimental variables impacting system performance.  Therefore,
we reserve the investigation on the impact of varying puck
densities for future work.

The role of dynamic task allocation in this domain requires the
robots to split their numbers by having some forage for
Puck\(_{Red}\) pucks and others for Puck\(_{Green}\) pucks.  For
the purpose of our experiments, we desire an allocation of robots
to converge to a situation in which the proportion of robots
foraging for Puck\(_{Red}\) pucks is equal to the proportion of
Puck\(_{Red}\) pucks present in the foraging arena (e.g., if
Puck\(_{Red}\) pucks make up 30\% of the pucks present in the
foraging arena, then 30\% of the robots should be foraging for
Puck\(_{Red}\) pucks).  In general, the desired allocation could
take other forms.  For example, it could be related to the
relative reward or cost of foraging each puck type without change
to our approach.

We note that the limited sensing capabilities and lack of direct communication of the individual robots in the implementation of our task domain prohibits them from acquiring global information such as the size and shape of the foraging arena, the initial or current number of pucks to be foraged (total or by type), or the initial or current number of foraging robots (total or by foraging type).

\subsection{Simulation Environment \label{simulation}}

In order to experimentally demonstrate the dynamic task allocation
mechanism we made use of a physically-realistic simulation
environment.  Our simulation trials were performed using Player
and Gazebo simulation environments.  Player~\cite{Player} is a
server that connects robots, sensors, and control programs over a
network.  Gazebo \cite{Gazebo} simulates a set of Player devices
in a 3-D physically-realistic world with full dynamics.  Together,
the two represent a high-fidelity simulation tool for individual
robots and teams that has been validated on a collection of
real-robot robot experiments using Player control programs
transferred directly to physical mobile robots.
\begin{figure}
\begin{center}
\includegraphics[height=1.7in]{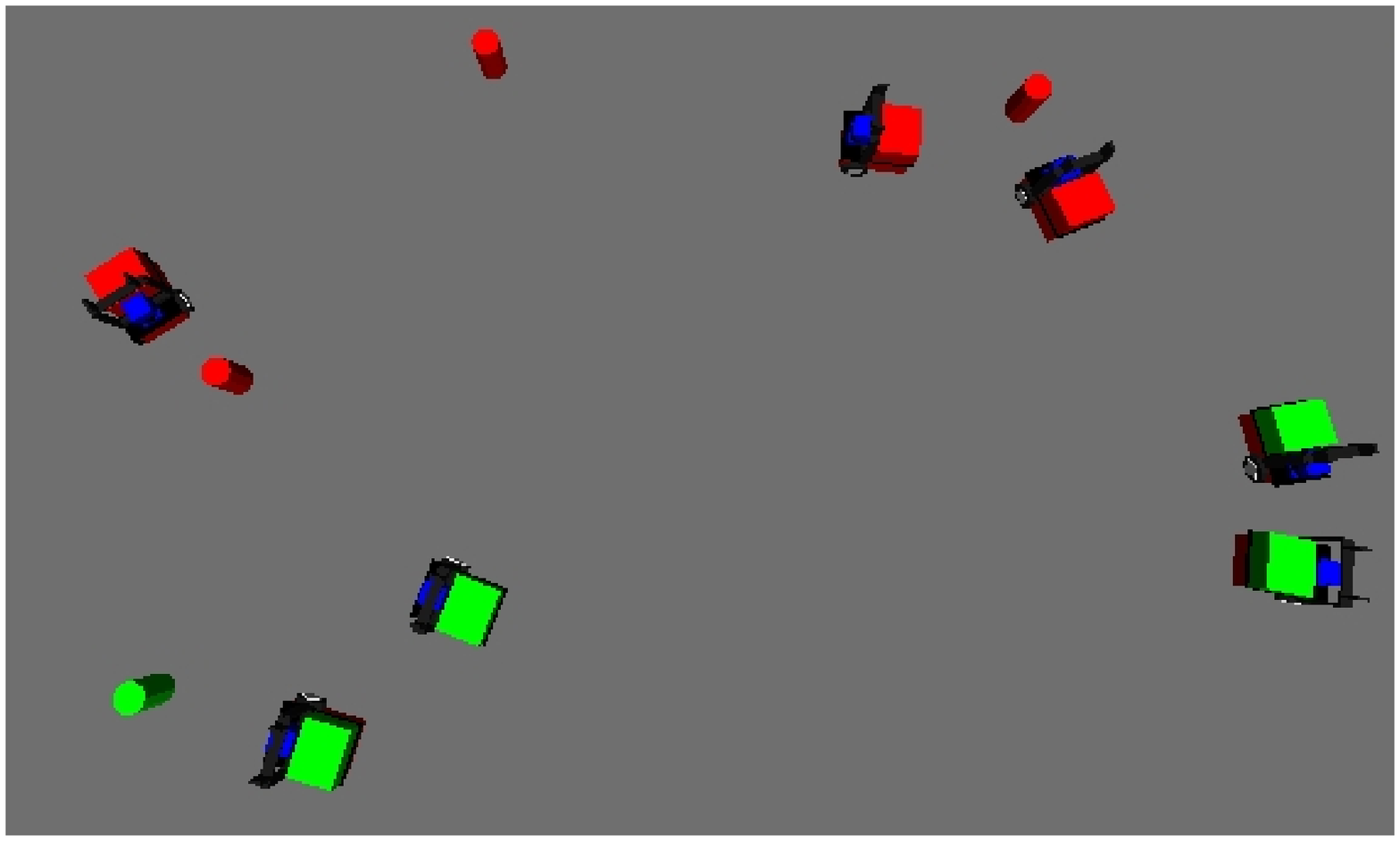}
\includegraphics[height=1.7in]{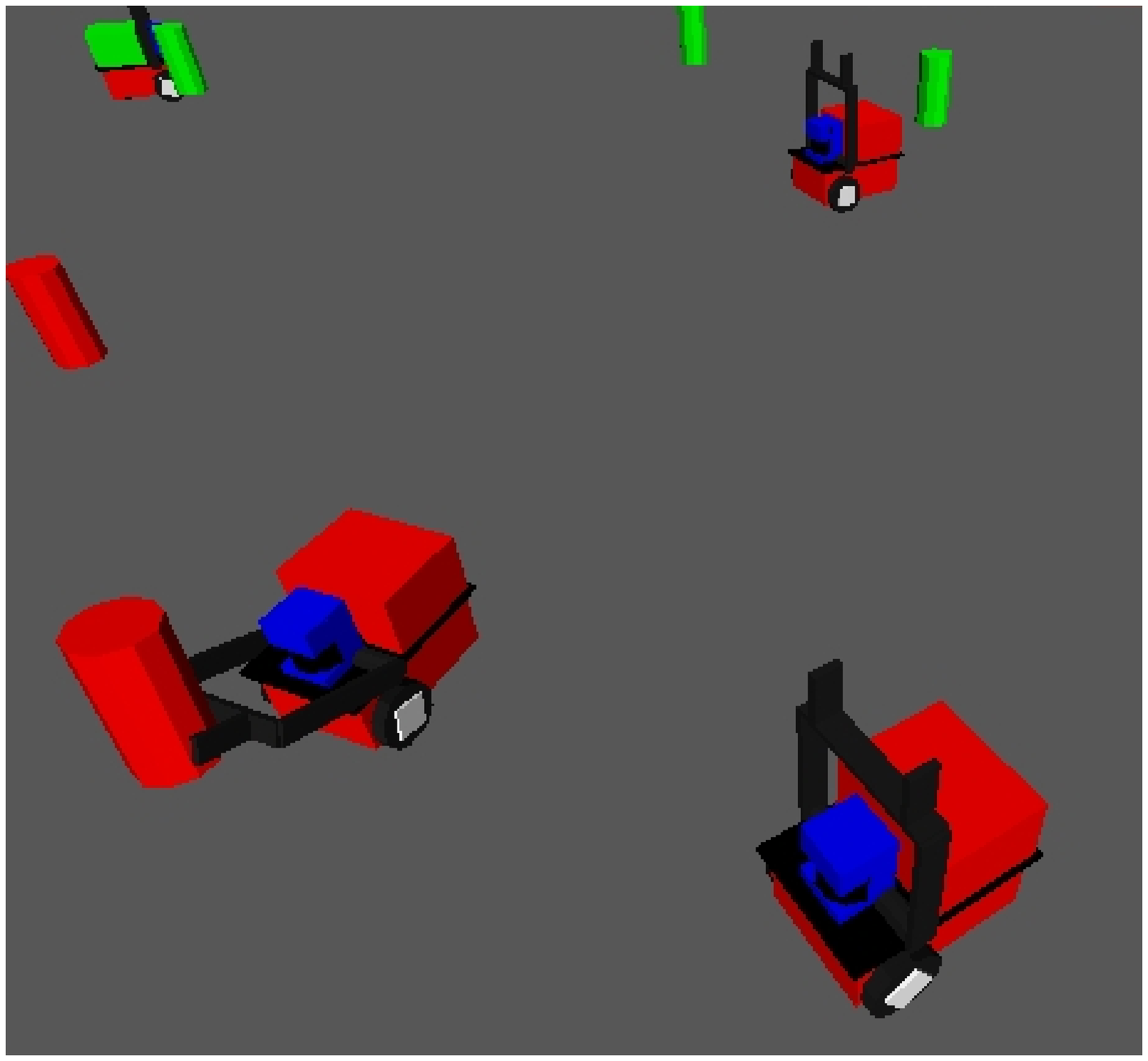}
\caption{Snapshots from the simulation environment used. (left) An overhead view of foraging arena and robots.  (right) A closeup of robots and pucks.} \label{fig:gazebo_snapshot}
\end{center}
\end{figure}
Figure~\ref{fig:gazebo_snapshot} provides snapshots of the simulation environment used.  All experiments involved 20 robots foraging in a 400m\(^2\) arena.

The robots used in the experimental simulations are realistic models of the ActivMedia Pioneer 2DX mobile robot.  Each robot, approximately 30 cm in diameter, is equipped with a differential drive, an odometry system using wheel rotation encoders, and 180 degree forward-facing laser rangefinder used for obstacle avoidance and as a fiducial detector/reader.  Each puck is marked with a fiducial that marks the puck type and each robot is equipped with a fiducial that marks the active foraging state of the robot.  Note that the fiducials do not contain unique identities of the pucks or robots but only mark the type of the puck or the puck type a given robot is engaged in foraging.  Each robot is also equipped with a 2-DOF gripper on the front, capable of picking up a single 8 cm diameter puck at a time.  There is no capability available for explicit, direct communication between robots nor can pucks and other robots be uniquely identified.

\subsection{Behavior-Based Robot Controller\label{controller}}

All robots have identical behavior-based controllers consisting of the following mutually exclusive behaviors: Avoiding, Wandering, Puck Servoing, Grasping, and Observing.  Descriptions of robot behaviors are provided below.

\begin{list}{-}{}
\item The {\bf Avoiding} behavior causes the robot to turn to avoid obstacles in its path.
\item The {\bf Wandering} behavior causes the robot to move forward and, after a random length of elapsed time, to turn left or right through a random arc for a random period of time.
\item The {\bf Puck Servoing} behavior causes the robot to move toward a detected puck of the desired type.  If the robot's current foraging state is Robot\(_{Red}\), the desired puck type is Puck\(_{Red}\), and if the robots current foraging state is Robot\(_{Green}\), the desired puck type is Puck\(_{Green}\).
\item The {\bf Grasping} behavior causes the robot to use its gripper to pick up and consume a puck within the gripper's grasp.
\item The {\bf Observing} behavior causes the robot to take the current fiducial information returned by the laser rangefinder and record the detected pucks and robots to their respective histories.  The robot then updates its foraging state based on those histories.  A description of the histories is given in \secref{state-info} and a description of the foraging state update procedure is given in \secref{transition-functions}.
\end{list}

\begin{table*}[t]
\begin{center}
\begin{tabular}{|c|c|c|c|c|}
\hline Obstacle & Puck\(_{Det}\) & Gripper Break- & Observation & Active \\
       Detected & Detected & Beam On & Signal & Behavior \\
\hline
\hline X & X & X & 1 & Observing \\
\hline 1 & X & X & X & Avoiding \\
\hline 0 & 1 & 0 & 0 & Puck Servoing \\
\hline 0 & X & 1 & 0 & Grasping \\
\hline 0 & X & X & X & Wandering \\ \hline
\end{tabular}
\end{center}
\caption{Behavior Activation Conditions.  Behaviors are listed in order of decreasing rank.  Higher ranking behaviors preempt lower ranking behaviors in the event multiple are active.  X denotes the activation condition is irrelevant for the behavior.  \label{beh-table}}
\end{table*}

Each behavior listed above has a set of activation conditions based on relevant sensor inputs and state values.  When met, the conditions cause the behavior to be become active.  A description of when each activation condition is active is given below.  The activation conditions of all behaviors are shown in Table~\ref{beh-table}.

\begin{list}{-}{}
\item The {\bf Obstacle Detected} activation condition is true when an obstacle is detected by the laser rangefinder within a distance of 1 meter.  Other robots, pucks, and the arena walls are considered obstacles.
\item The {\bf Puck\(_{Det}\) Detected} activation condition is true if the robot's current foraging state is Robot\(_{Det}\) and a puck of type Puck\(_{Det}\) (where Det is {\it Red} or {\it Green}) is detected within a distance of 5 meters and within \(\pm\) 30 degrees of the robot's direction of travel.
\item The {\bf Gripper Break-Beam On} activation condition is true if the break-beam sensor between the gripper jaws detects an object.
\item The {\bf Observation Signal} activation condition is true if the distance traveled by the robot according to odometry since the last time the {\bf Observing} behavior was activated is greater than 2 meters.
\end{list}

\subsubsection{Robot State Information \label{state-info}}

All robots maintain three types of state information: foraging state, observed puck history, and observed robot history.  The foraging state identifies the type of puck the robot is currently involved in foraging.  A robot with a foraging state of Robot\(_{Red}\) refers to a robot engaged in foraging Puck\(_{Red}\) pucks and a foraging state of Robot\(_{Green}\) refers to a robot engaged in foraging Puck\(_{Green}\) pucks.  For simplicity, we will refer to both robot foraging states and puck types as $Red$ and $Green$. The exact meaning will be clear in context.

Each robot is outfitted with a colored beacon passively observable by nearby robots which indicates the robot's current foraging state.  The color of the beacon changes to reflect the current state -- a red beacon for a foraging state of $Red$ and a green beacon for foraging state $Green$.  Thus, the colored beacon acts as a form of local, passive communication conveying the robot's current foraging state.  All robots maintain a limited, constant-sized history storing the most recently observed puck types and another constant-sized history storing the foraging state of the most recently observed robots.  Neither of these histories contains a unique identity or location of detected pucks or robots, nor does it store a time stamp of when any given observation was made.  The history of observed pucks is limited to the last {\tt MAX-PUCK-HISTORY} pucks observed and the history of the foraging states of observed robots is limited to the last {\tt MAX-ROBOT-HISTORY} robots observed.

While moving about the arena, each robot keeps track of the approximate distance it has traveled by using odometry measurements.  At every interval of 2 meters traveled, the robot makes an observation performed by the {\it Observing} behavior.  This procedure is nearly instantaneous; therefore, the robot's behavior is not outwardly affected.  The area in which pucks and other robots are visible is within 5 meters and \(\pm\) 30 degrees in the robot's direction of travel.  Observations are only made after traveling 2 meters because updating too frequently leads to over-convergence of the estimated puck and robot type proportions due to repeated observations of the same pucks and/or robots.  On average, during our experiments, a robot detected 2 pucks and robots  per observation.

\subsubsection{Foraging State Transition Function \label{transition-functions}}

After a robot makes an observation, it re-evaluates and probabilistically changes its current foraging state given the newly updated puck and robot histories. The probability by which the robot changes its foraging state is defined by the transition function.  We experimentally studied transition functions given by \eqref{eq:rates}, \eqref{eq:fR} and \eqref{eq:fG} with both $power$ and $linear$ forms. Below we present results of analysis and simulations and discuss the consequences the choice of the transition function has on system level behavior.

%%%%%%%%%%%%%%%%%%pucks only results%%%%%%%%%%%%%%%%%%%%%%%%%%%%%%%%%%
\section{Analysis and Simulations Results}
\label{sec:results}
The mathematical models developed in \secref{sec:analysis} can be directly applied to the multi-foraging task if we map $Red$ and $Green$ tasks to $Red$ and $Green$ pucks and task states of robots to their foraging states. Model parameters, such as $\varepsilon$, $\alpha$, etc, depend on physical realizations of the implementation and can be computed from details of the multi-foraging task as described below.

\subsection{Observations of Pucks Only}
\label{sec:results1} First, we study the model of dynamic task
allocation, presented in \secref{sec:pucksonly}, where robots
observe only pucks and make decision to switch foraging state
according to the transition functions given by \eqref{eq:rates}.
We compared theoretical predictions of the robots' collective
behavior with results from simulations. We used
\eqref{eq:solution} and \ref{eq:jump} to compute how the average
number of robots in the $Red$ state changes in time when the puck
distribution is suddenly changed. The parameter values were
obtained from experiments. $p_0 = 1.0$ was the initial density of
$Red$ robots (of $20$ total robots), $\mu_0=0.3$ was the initial
$Red$ puck density (of $50$ total pucks), which remained constant
until it was changed by the experimenter. The first change in puck
density was $\Delta \mu=0.5$, meaning that $80\%$ of the pucks in
the arena are now $Red$. The second change in puck density was
$\Delta \mu=-0.3$, to $50\%$ $Red$ pucks.

$\epsilon$ is the rate at which robots make decisions to switch
states. Robot traveled $2\ m$ between observations at an average
speed of $0.2\ m/s$; therefore, there are $10\ s$ between
observations, and $\varepsilon=0.1$. $h$, the history length, is
the number of pucks in the robot's memory. $\alpha M^0$ is the
rate at which robots encounter pucks. \commentout{
% continuous observations
As a robot travels through the arena, it sweeps out some area
during time $dt$ and will detect objects that fall in that area.
This detection area is $V_R W_R dt$, where $V_R=0.2\ m/s$ is
robot's average speed, and $W_R=5\ m$ is robot's detection width.
The area of the arena is $A=315\ m^2$; therefore, a robot will
detect pucks at a rate $\alpha = V_R W_R /A  = 0.003\ s^{-1}$. } A
robot makes an observation of its local environment at discrete
time intervals. The area visible to the robot is $A_{vis}=(5\ m)^2
\pi/6 =13.09$, with $1/6$ coming from the $60^o$ angle of view.
The arena area is $A=315\ m^2$; therefore,  $\alpha
M^0=A_{vis}M^0/A=2.1$. We studied the dynamics of the system for
different history lengths $h$.

\begin{figure}[tbhp]
\includegraphics[width=1.0\textwidth]{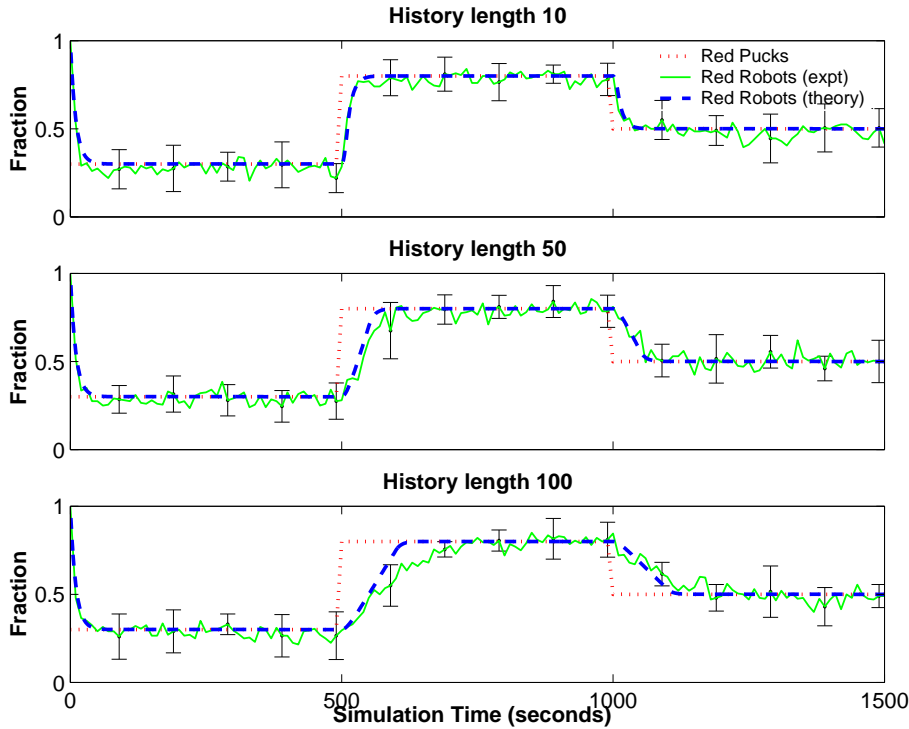}
\caption{Evolution of the fraction of $Red$ robots for different
history lengths. Robots' decision to change state is based on observations of pucks only.} \label{fig:puckonly}
\end{figure}

\figref{fig:puckonly} shows evolution of the numbers of $Red$
robots for different history lengths. Initially, the distribution
of $Red$ pucks is set to $30\%$ and all the robots are in the
$Red$ foraging state. At $t=500\ s$, the puck distribution changes
abruptly to $80\%$, and at $t=1000\ s$ to $50\%$. The solid line
shows results of simulations --- the fraction of $Red$ robots,
averaged over 10 runs. The dashed line gives theoretical
predictions for the parameters quoted above.
 Since we are in the $\varepsilon h \gg
1$ limit (for $h=50,100$), the time it takes to converge to the
steady state is linear in history length, $t_{conv}\sim h$, as
predicted by \eqref{eq:jump}. The agreement between theoretical
and experimental results is excellent. We stress that there are no
free parameters in the theoretical predictions --- only
experimental values of the parameters were used in producing these
plots.

In addition to being able to predict the average collective
behavior of the multi-robot system, we can also quantitatively
characterize the amount of fluctuations in the system.
Fluctuations are deviations from the steady state (after the
system has converged to the steady state) that arise from the
stochastic nature of robot's observations and decisions.
  These deviations result in fluctuations
from the desired global distribution of $Red$ and $Green$ robots
seen in an individual experiment. One can suppress these
fluctuations by averaging results of many identical experiments.

\begin{figure}[tbhp]
\center{
\includegraphics[width=0.5\textwidth]{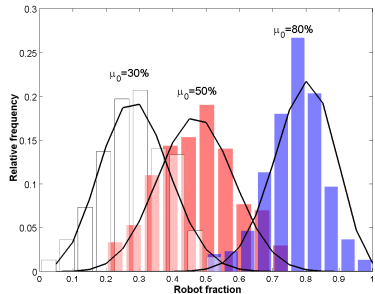}
\caption{Histogram of the fraction of $Red$ robots in the steady
state for three different puck distributions (data for $h=10$).
$\mu_0$ specifies fraction of $Red$ pucks. Lines are theoretical
predictions of the distribution of $Red$ robots.} }
\label{fig:fluctuations}
\end{figure}

To measure the strength of the fluctuations, we take data from an
individual experimental run and extract the fraction of $Red$
robots, after the system has converged to the steady state, for
each of the three $Red$ puck distributions: $\mu_0=30\%,\ 50\%,\
80\%$. Because the runs were relatively short, we only have $300\
s$ worth of data (30 data points) in the converged state; however,
since each experiment was repeated ten times, we make the data
sets longer by appending data from all experiments. In the end, we
have 300 measurements of the steady state $Red$ robot density for
three different puck distributions. \figref{fig:fluctuations}
shows the histogram of robot distributions for three different
puck distributions. The solid lines are computed using
\eqref{eq:Pn}, where for $\overline{\gamma}$ we used the actual
means of the steady state distributions ($\overline{\gamma}=0.28$,
$0.47$ and $0.7$ for $\mu_0=30\%$, $50\%$ and $80\%$
respectively). We can see from the plots that the theory correctly
predicts the strength of fluctuations about the steady state. As
is true of binomial distributions, the fluctuations (measured by
the variance) are greatest for cases where the numbers of $Red$
and $Green$ pucks are comparable ($\mu_0=50\%$) and smaller when
their numbers are very different ($\mu_0=80\%$).

\subsection{Observations of Pucks and Robots}
\label{sec:results2} In this section we study the dynamic task
allocation model developed in \secref{sec:phenomenological}, in
which robots use observations of pucks and other robots' foraging
states to make decision to change their own foraging state.

\begin{figure}[tbhp]
\begin{tabular}{c}
\includegraphics[width=0.9\textwidth]{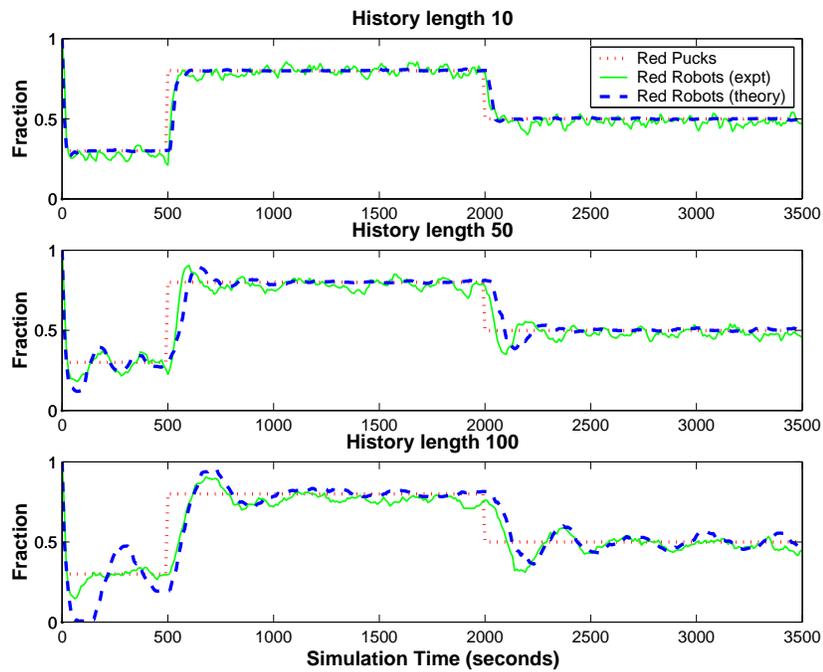}
\\
(a) Linear transition function
\\
\includegraphics[width=0.9\textwidth]{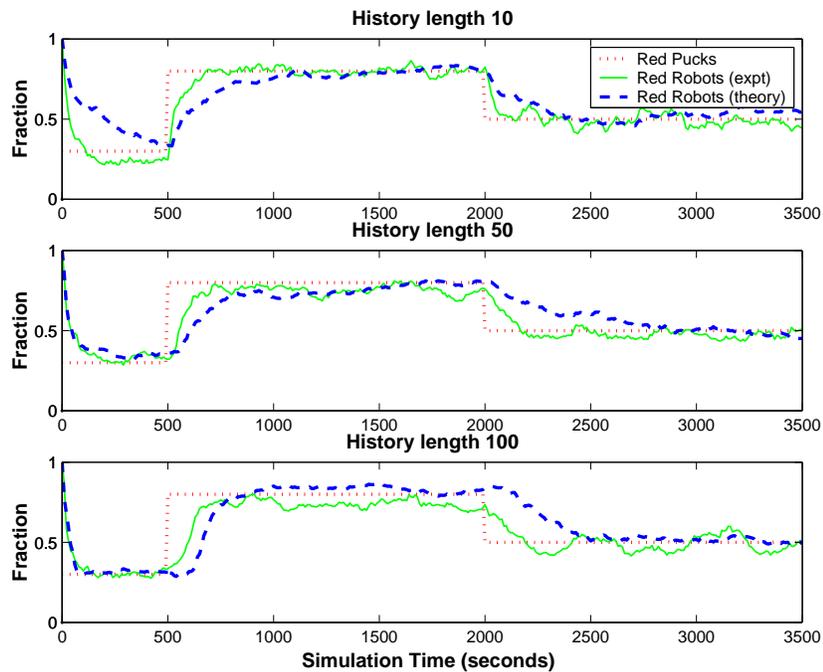}
\\
(b) Power transition function
\end{tabular}
 \caption{Evolution of the fraction of $Red$ robots for
different history lengths and transition functions, compared to
predictions of the model} \label{fig:noise}
\end{figure}

\figref{fig:noise} shows results of embodied simulations (solid
lines) as well as solutions to the model \eqref{eq:stochastic} (dashed lines) for
different values of robot history length and forms of transition
function (given by \eqref{eq:fR} and \ref{eq:fG}, with $g(z)$ linear or power function). Initially, the $Red$ puck fraction (dotted line) is
$30\%$. It is changed abruptly at $t=500\ s$ to $80\%$ and then
again at $t=2000\ s$ to $50\%$. Each solid line showing $Red$
robot density has been averaged over 10 runs. We rescale the dimensionless time of the model by
parameter $10$, corresponding $\varepsilon=0.1$. The history
length was the only adjustable parameter used in solving the equations.
The values of $h$ used to compute the observed fraction of $Red$
robots $n_r$ in \eqref{eq:2} were $h=2,\ 8,\ 16$, corresponding
to experimental history lengths $10,\ 50,\ 100$ respectively. For
$m_r$, the observed fraction of $Red$ pucks, we used their actual
densities.

In order to explain the difference in history lengths between
theory and experiment, we note that in the simulation experiments, the history
length means the numbers of observed robots and pucks, while in
the model, it means the number of observations, with multiple
objects sighted within a single observation. According to
calculations in \secref{sec:results1}, a robot observes about 2
pucks in a single observation. Moreover, the robot travels $2\ m$
between observations, yet it sees $5\ m$ out during each
observation, meaning that individual observations will be
correlated. Observations will be further correlated because of the
pattern of a robot's motion --- as the robot moves in a straight
line towards a goal, it is likely to observe overlapping regions of
the arena. These considerations could explain the factor of five difference between the history lengths used in the
experiments and the corresponding values used in the model. More
detailed experiments, for example, ones in which robots travel
farther between observations, are necessary to explain these
differences.

Solutions exhibit oscillations, although eventually oscillations
decay and solutions relax to their steady state values. In all
cases, the steady state value is the same as the fraction of red
pucks in the arena. History-induced oscillations are far more
pronounced for the linear transition function
(\figref{fig:noise}(a)) than for the power transition function
(\figref{fig:noise}(b)). For the power transition function, these
oscillations are present but become evident only for longer
history lengths. This behavior is probably caused by the
differences between the values of transition functions near the
steady state: while the value of the power transition function
remains small near the steady state, the value of the linear
transition function grows linearly with the distance from the
steady state, thereby amplifying any deviations from the steady
state solution. The amplitude and period of oscillations and the
convergence rate of solutions to the steady state all depend on
history length, and it generally takes longer to reach the steady
state for longer histories. Another conclusion is that the linear
transition function converges to the desired distribution faster
than the power function, at least for moderate history lengths.

\section{Discussion}
\label{sec:discussion}

We have constructed and analyzed mathematical models of dynamic
task allocation in a multi-robot system. The models are general
and can be easily extended to other systems in which robots use a
history of local observations of the environment as a basis for
making decisions about future actions. These models are based on
theory of stochastic processes. In order to study a robot's
behavior, we do not need to know its exact trajectory or the
trajectories of other robots; instead, we derive a probabilistic
model that governs how a robot's behavior changes in time. In some
simple cases these models can be solved analytically. However,
stochastic models are usually too complex for exact analytic
treatment. Thus, in the scenario described in
\secref{sec:pucksonly} in which only observations of tasks are
made, though the individual model is tractable, the stochastic
model of the collective behavior is not. Instead, we use averaging
and approximation techniques to quantitatively study the dynamics
of the collective behavior. Such models, therefore, do not
describe the robots' behavior in a single experiment, but rather
the behavior that has been averaged over many experimental or
simulations runs. Fortunately, results of experiments and
simulations are usually presented as an average over many runs;
therefore, mathematical models of average collective behavior can
be used to describe experimental results. In fact, the stochastic
model produces excellent agreement with experimental results under
all experimental conditions and without using any adjustable
parameters.

Phenomenological models are more straightforward to construct and
analyze than exact stochastic models --- in fact, they can be
easily constructed from details of the individual robot
controller~\cite{Lerman04sab}. The ease of use comes at a price,
namely, the number of simplifying assumptions that were made in
order to produce a mathematically tractable model. First, we
assume that the robots are functioning in a dilute limit, where
they are sufficiently separated that their actions are largely
independent of one another. Second, we assume that the transition
rates can be represented by aggregate quantities that are
spatially uniform and independent of the details of the individual
robot's actions or history. We also assume the system is
homogeneous, with modeled robots characterized by a set of
parameters, each of them representing the mean value of some real
robot feature: mean speed, mean duration for performing a certain
maneuver, and so on. Real robot systems are heterogeneous: even if
the robots are executing the same controller, there will always be
variations due to inherent differences in hardware. We do not
consider parameter distributions in our models as would be
necessary to describe such heterogeneous systems. Finally,
phenomenological models more reliably describe systems where
fluctuations (deviations from the mean behavior) can be neglected,
as happens in large systems or when many experimental runs are
aggregated. However, even if phenomenological models don't agree
with experiments exactly, as we saw in \secref{sec:results2}, they
can still reliably predict most behaviors of interest even in
not-so-large systems. They are, therefore, a useful tool for
modeling and analyzing multi-robot systems.

\section{Conclusion}
\label{sec:conclusion}

 Mathematical analysis can be a useful tool
for the study and design of MRS and  a viable alternative to
experiments and simulations. It can be applied to large systems
that are too costly to build or take too long to run in
simulation. Mathematical analysis can be used to study the
behavior of an MRS, select parameters that optimize its
performance, prevent instabilities, \emph{etc}. In conjunction
with the design process, mathematical analysis can help understand
the effect individual robot characteristics have on the collective
behavior \emph{before} a system is implemented in hardware or in
simulation. Unlike experiments and simulations, where exhaustive
search of the design parameter space is often required to reach
any conclusion, analysis can often produce exact analytic results,
or scaling relationships, for the quantities of interest. If these
are not possible, exhaustive search of the parameter space is much
more practical and efficient. Finally, results of analysis can be
used as feedback to guide performance-enhancing modifications of
the robot controller.

In this paper we have described an dynamic task allocation
mechanism where robots use local observations of the environment
to decide their task assignments. We have presented a mathematical
model of this task allocation mechanism and studied it in the
context of a multi-foraging task scenario. We compared predictions
of the model with results of embodied simulations and found
excellent quantitative agreement. In this application,
mathematical analysis could help the designer choose robot
properties, such as the form of the transition probability used by
robots to switch their task state, or decide how many observations
the robot ought to consider.

Mathematical analysis of MRS is a new field, but its success in
explaining experimental results shows it to be a promising tool
for the design and analysis of robotic systems. The field is open
to new research directions, from applying analysis to new robotic
systems to developing increasingly sophisticated mathematical
models that, for example, account for heterogeneities in robot
population that are due to differences in their sensors and
actuators.

\section*{Acknowledgment}
The research reported here was supported in part by the Defense Advanced Research Projects Agency (DARPA) under contract number F30602-00-2-0573.

\bibliographystyle{plain}
% argument is your BibTeX string definitions and bibliography database(s)
%\bibliography{IEEEabrv,../bib/paper}
\bibliography{../../../tex/bib/agents,../../../tex/bib/robots,../../../tex/bib/lerman,../../../tex/bib/physics}

\end{document}